\makeatletter
\def\input@path{{IEEE-Transactions-templates-and-instructions/}}
\makeatother
\documentclass[letterpaper,journal]{IEEEtran}

\usepackage[T1]{fontenc}
\usepackage{amsmath,amssymb,amsfonts}
\usepackage{array}
\usepackage{booktabs}
\usepackage{cite}
\usepackage{graphicx}
\usepackage{multirow}
\usepackage{placeins}
\usepackage{stfloats}
\usepackage{textcomp}
\usepackage{url}
\usepackage{verbatim}
\usepackage{xcolor}
\usepackage{hyperref}
\usepackage{cleveref}

\hypersetup{hidelinks}
\crefname{figure}{Fig.}{Figs.}
\Crefname{figure}{Fig.}{Figs.}
\crefname{table}{Table}{Tables}
\Crefname{table}{Table}{Tables}

\newcommand{\ntrials}{10}
\newcommand{\npyramidtrials}{5}
\newcommand{\method}{ActionReasoning~2.0}

\begin{document}
\bstctlcite{BSTcontrol}

\title{A Brain-inspired Hierarchical Framework for Zero-Shot Robot Task Reasoning and Execution}

\author{\mbox{Guangming Wang}, \mbox{Pengfei Ye}, \mbox{Qizhen Ying}, \mbox{Yixiong Jing}, \mbox{Yuxiang Ma}, \mbox{Haonan Chen}, \mbox{Haibing Wu}, \mbox{Olaf Wysocki}, \mbox{Molong Duan}, and \mbox{Brian Sheil}%
\thanks{G. Wang, Y. Jing, H. Wu, O. Wysocki, and B. Sheil are with the Department of Engineering, University of Cambridge, Cambridge, U.K.}%
\thanks{P. Ye is with the Computer Science and Artificial Intelligence Laboratory, Massachusetts Institute of Technology, Cambridge, MA, USA, and also with the Department of Mechanical and Aerospace Engineering, Hong Kong University of Science and Technology, Hong Kong, China.}%
\thanks{Q. Ying is with the Department of Engineering Science, University of Oxford,
Oxford, U.K.}
\thanks{Y. Ma is with the Computer Science and Artificial Intelligence Laboratory, Massachusetts Institute of Technology, Cambridge, MA, USA.}%
\thanks{H. Chen is with Computer Science and Kempner Institute, Harvard University, Cambridge, MA, USA.}%
\thanks{M. Duan is with the Department of Mechanical and Aerospace Engineering, Hong Kong University of Science and Technology, Hong Kong, China.}%
\thanks{G. Wang and P. Ye contributed equally. Corresponding authors: Yixiong Jing and Molong Duan (e-mail: yj401@cam.ac.uk and duan@ust.hk).}}

\maketitle

\begin{abstract}
Robots that follow open-ended language instructions need to connect semantic intent to visual scene understanding, geometric feasibility, object states, and physical interaction conditions. End-to-end Vision-Language-Action policies have improved cross-task generalization, but they typically map visual and language inputs directly to robot actions, leaving limited explicit structure for long-horizon decomposition, physical verification, and recovery. We present \method, a zero-shot hierarchical framework functionally inspired by the division of roles in the human brain, comprising visual perception and state inference, language grounding and action-sequence generation from a shared atomic action library, cost-based plan selection, and real-robot execution and verification. The framework grounds commands in explicit object states, composes reusable atomic actions into task-conditioned sequences, ranks alternative sequences by execution cost, and verifies intermediate physical outcomes from refreshed observations. In the evaluation, \method{} completes 10/10 clean board trials, 10/10 pick-and-place trials, and 4/5 pyramid stacking trials for both the flat and irregular initial-layout conditions; the corresponding mean task progress is $99.03\%$, $100.00\%$, and $96.67\%$ respectively. Across all evaluated conditions, \method{} achieves higher success rates than ReKep, Dream2Flow, and $\pi_{0.5}$ benchmarks, demonstrating the effectiveness of combining explicit object-state reasoning, compositional atomic actions, cost-based plan selection, and closed-loop execution verification.
\end{abstract}


\section{Introduction}

\IEEEPARstart{N}{ew} robotic applications increasingly require autonomous execution of complex and long-horizon manipulation tasks \cite{yao2025longhorizon,liu2023motionplanning}. Recent progress in Large Language Models (LLMs) and Vision-Language Models (VLMs) has further driven the emergence of Vision-Language-Action (VLA) models for language-conditioned embodied decision-making and action generation \cite{ma2024survey,brohan2023rt2,kim2024openvla,black2025pi05}. However, reliable long-horizon execution in the physical world also requires robot actions to remain consistent with physical constraints such as geometry, object states, and interaction conditions. Structured planning and grounding approaches address this requirement by explicitly incorporating motion feasibility, relational constraints, or interaction constraints \cite{garrett2021tamp,huang2024rekep,pan2025omnimanip}, whereas generalist VLA policies typically do not provide an explicit mechanism for enforcing such constraints during execution. Reliable long-horizon execution therefore calls for feasible action sequences together with verification of intermediate physical states. Commands such as "build a pyramid", "put all blocks in the bowl", and "clean the whiteboard" correspond to complex sequences of individual robot actions, many of which recur across different tasks. We refer to these reusable elementary operations as \emph{atomic actions}, which can be composed into task-conditioned action sequences for complex robot execution. 
A practical gap therefore remains in how to retain the flexibility and generalization capability of foundation models while achieving robust and adaptive physical execution across different robot tasks. Bridging this gap calls for a more structured organization of reasoning and execution, motivating us to consider how complementary functional roles can be coordinated within a unified robotic intelligence framework.

\begin{figure*}[!t]
    \centering
    \includegraphics[width=1.0\linewidth]{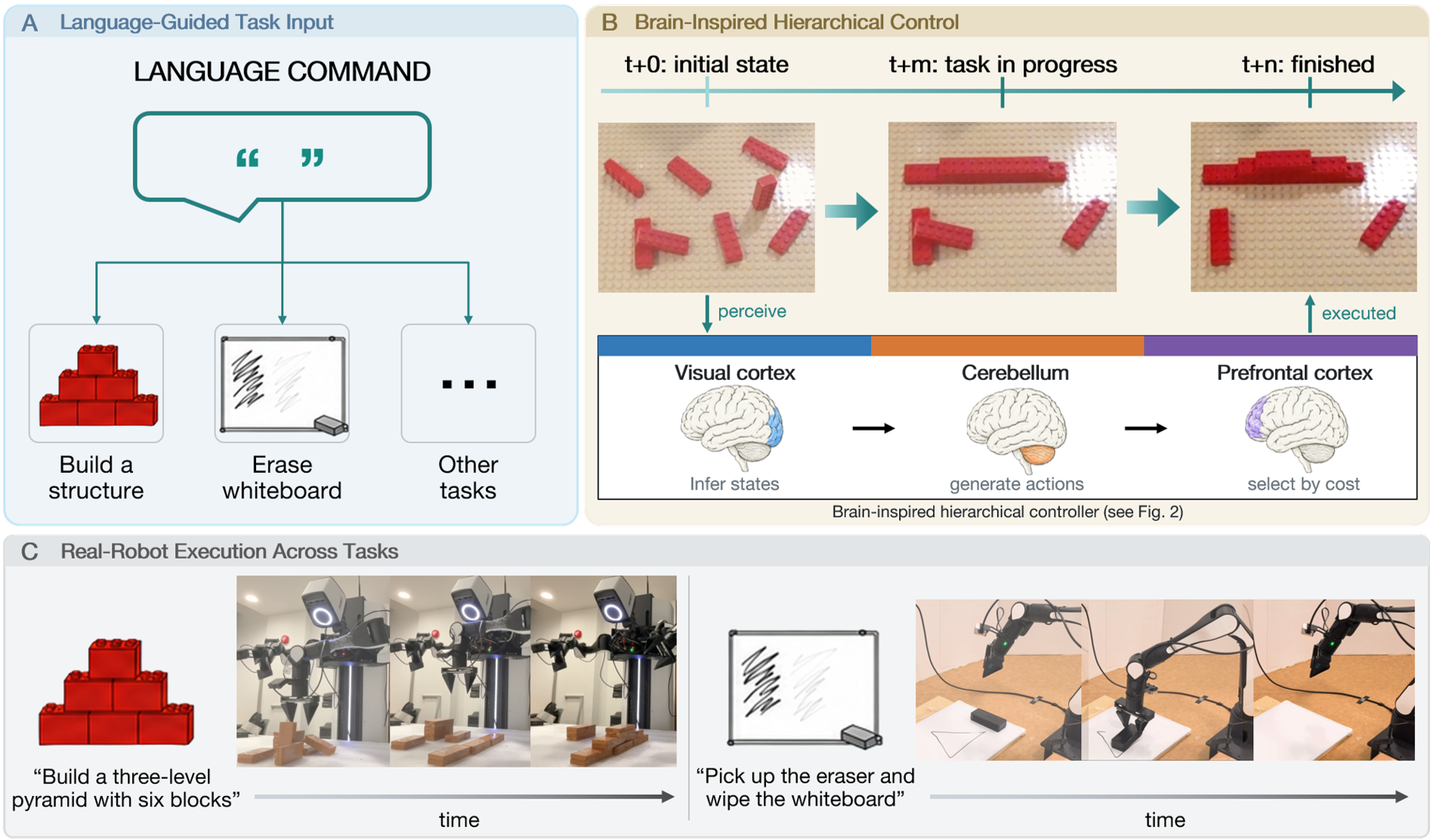}
    \caption{Overview of \method{}. (A) A natural-language command specifies the robot task. (B) A brain-inspired hierarchy organizes perception and state inference, language grounding and atomic-action composition, and cost-based plan selection through functional analogies to the visual cortex, cerebellum, and prefrontal cortex. (C) The resulting task plans are executed on real robots across construction and surface-cleaning tasks, with updated observations supporting verification until task completion.}
    \label{fig:framework}
\end{figure*}

In this paper, we introduce \method, a hierarchical reasoning framework for zero-shot robot task execution. As illustrated in \cref{fig:framework}, \method{} combines a natural-language task command with visual observations of the workspace and organizes reasoning through a functional analogy to human brain organization. A visual-cortex-inspired module performs perception and state inference; a cerebellum-inspired module grounds the command and composes atomic actions from the shared library into candidate action sequences; and a prefrontal-cortex-inspired module evaluates and ranks these sequences using an execution cost that favors shorter motions and fewer additional operations. The selected task plan is then executed on the robot with intermediate-state verification.

We evaluate \method{} on three complementary task families: pick-and-place, which tests repeated object selection and placement; clean board, which tests continuous-contact manipulation and coverage; and pyramid stacking, which requires long-horizon construction and stability reasoning under both flat and irregular initial layouts.

Our main contributions can be summarised as follows:
\begin{itemize}
    \item We introduce a brain-inspired hierarchical framework for zero-shot robot task execution that integrates object-state perception, language-grounded composition of atomic actions into action sequences, cost-based plan selection, and execution verification.
    \item We develop a unified action-sequence generation formulation built on a shared atomic action library, with cost-based plan ranking and task-specific execution verification, enabling the same reasoning pipeline to support pick-and-place, stacking, and surface-cleaning tasks.
\item We benchmark \method{} against ReKep, Dream2Flow, and $\pi_{0.5}$ across three task families, achieving the highest success rate in every evaluated condition, and validate the effectiveness of key components through ablation studies.
\end{itemize}

\section{Related Work}
\label{sec:related-work}

\subsection{Generalist Robot Policies and Action Representations}

Robot learning has increasingly shifted from task-specific policies toward language-conditioned generalist models. RT-1 discretizes continuous robot actions into tokens, while RT-2 represents robot actions as text tokens within a vision-language-action framework \cite{brohan2022rt1,brohan2023rt2}. OpenVLA and $\pi_{0.5}$ further develop generalist VLA policies trained on diverse vision-language and robot data \cite{kim2024openvla,black2025pi05}. Beyond autoregressive action prediction, diffusion policy formulates continuous action generation as a conditional denoising process, while FAST introduces an efficient frequency-domain representation for tokenizing robot action sequences \cite{chi2023diffusionpolicy,pertsch2025fast}. These methods demonstrate increasingly flexible policy and action representations. In contrast, our framework keeps intermediate action sequences explicit and evaluates them before execution through cost-based plan selection and execution verification.

The breadth of generalist robot policies also depends strongly on the diversity of training data. Large-scale datasets such as BridgeData V2, DROID, and Open X-Embodiment substantially broaden the range of tasks, scenes, and robot embodiments available for policy learning \cite{walke2023bridgedata,khazatsky2024droid,openxembodiment2023rtx}. Octo further demonstrates generalist policy learning across heterogeneous robot datasets and embodiments \cite{octo2024}. While such data diversity improves generalization, it does not by itself provide an explicit mechanism for evaluating the physical feasibility of generated actions or verifying intermediate execution states.

\subsection{Structured Visuomotor Grounding}

Structured visuomotor methods provide complementary mechanisms for grounding perception, language, and robot actions. VIMA formulates manipulation through multimodal prompts, CLIPort combines semantic "what'' and spatial "where'' pathways, and PerAct predicts discretized 6-DoF voxel actions for manipulation \cite{jiang2022vima,shridhar2022cliport,shridhar2023peract}. OmniManip further represents object-centric interaction primitives as spatial constraints for general robotic manipulation \cite{pan2025omnimanip}. These approaches demonstrate the value of structured intermediate representations for connecting perception with action. Our framework instead uses explicit object states as an intermediate representation for language-grounded action-sequence generation, followed by cost-based plan selection and execution verification.

\subsection{Physics-Aware Planning and LLM Reasoning}

Task-and-motion planning integrates discrete task planning with continuous motion feasibility \cite{garrett2021tamp}. Recent language-model-based approaches introduce complementary mechanisms for grounding high-level reasoning in robot execution. SayCan ranks pretrained skills by combining language-model scores with learned affordance values, while Code as Policies synthesizes executable robot policy code from language instructions \cite{ahn2022saycan,liang2022codeaspolicies}. VoxPoser constructs composable 3-D value maps for language-guided manipulation, ReKep formulates manipulation through relational keypoint constraints, and Dream2Flow derives 3-D object flow from generated videos to guide manipulation \cite{huang2023voxposer,huang2024rekep,dharmarajan2025dream2flow}. Building on ActionReasoning, which introduced physics-aware 3-D action reasoning for robotic brick stacking \cite{wang2026actionreasoning}, we extend this direction to a unified hierarchy that integrates object-state perception, language-grounded action-sequence generation, cost-based plan selection, and execution verification across multiple task families.

Chain-of-thought reasoning demonstrates the value of explicit intermediate reasoning, while ReAct interleaves reasoning with actions to support interactive decision-making \cite{wei2022cot,yao2022react}. For robotic execution, however, semantic plausibility alone is insufficient; generated plans must also remain consistent with the observed physical state and execution constraints. \method{} therefore separates language-grounded action generation from cost-based plan evaluation and execution verification, allowing candidate action sequences to be explicitly assessed before and during real-robot execution.

\begin{figure*}[!t]
    \centering
    \includegraphics[width=1.0\linewidth]{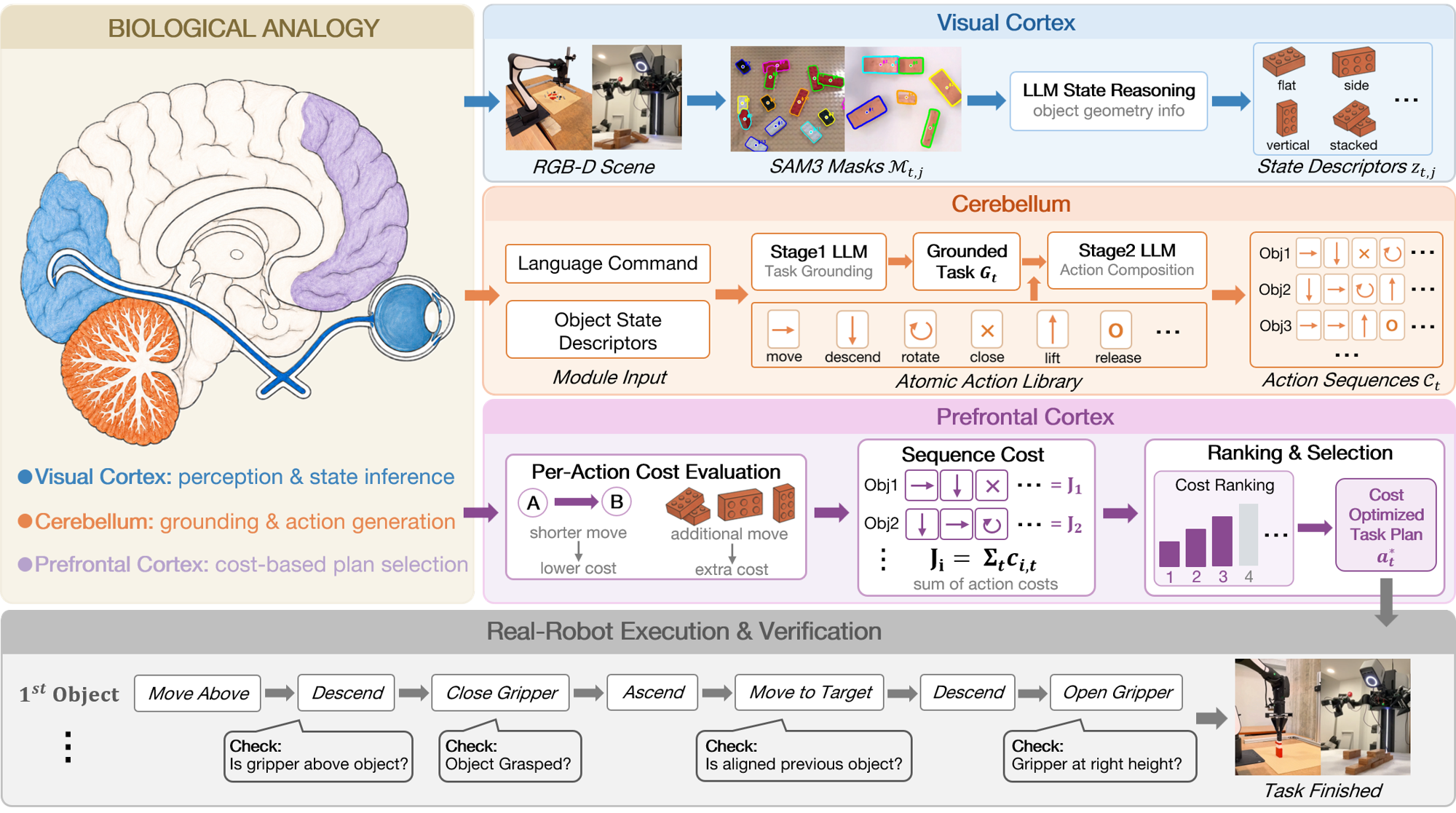}
    \caption{Brick-construction data flow. RGB-D and language inputs feed object-state, action-generation, and decision roles. Unique assignments are dispatched directly, surplus alternatives are optimized, and placement ends with a checked \textsc{Release}.}
    \label{fig:detailed_architecture}
\end{figure*}

\section{Methodology}
\label{sec:methodology}

\method{} is a hierarchical framework that maps a language command, RGB-D observations, and robot state into structured scene and task representations for physically verified robot execution. As illustrated in \cref{fig:detailed_architecture}, its organization is functionally inspired by the division of roles in the human brain: a visual-cortex-inspired module performs perception and state inference, a cerebellum-inspired module grounds the task and composes atomic actions into candidate action sequences, and a prefrontal-cortex-inspired module evaluates and ranks these sequences by execution cost. The selected plan is then passed to the real-robot execution and verification module. The analogy is functional rather than neurobiological.

The same hierarchy is used across all task families. \Cref{fig:detailed_architecture} details the data flow for brick construction, while pick-and-place and clean board retain the same interfaces and shared atomic action library, with task-specific goals, action-sequence composition, and verification conditions. During execution, updated observations is used to revise task progress and generate a new action sequence when required.

\subsection{Visual Perception and State Inference}
\label{subsec:perception-goals}

At execution step $t$, the robot receives an RGB-D observation $O_t$ and its current robot state $q_t$. As illustrated in \cref{fig:detailed_architecture}, the visual-perception module first extracts task-relevant objects and regions from the RGB-D scene and then infers their physical states. SAM~3 \cite{carion2025sam3} provides concept-conditioned segmentation masks, while registered depth projects each mask into the robot coordinate frame to recover the object position, orientation, and geometric extent.

We represent the resulting scene state as
\begin{equation}
S_t=\Phi(O_t,q_t)
=\left(\mathcal D_t,\mathcal R_t,q_t\right),
\label{eq:scene_state}
\end{equation}
where $\mathcal D_t=\{d_{t,j}\}_{j=1}^{N_t}$ denotes the set of object-state descriptors, $\mathcal R_t$ represents spatial relations among task-relevant entities, and $q_t$ denotes the current robot state. Each object-state descriptor is defined as
\begin{equation}
d_{t,j}
=
\left(
\omega_j,
\mathcal M_{t,j},
\mathbf T_{t,j},
\mathbf b_{t,j},
z_{t,j}
\right),
\label{eq:object_descriptor}
\end{equation}
where $\omega_j$ denotes the entity identity, $\mathcal M_{t,j}$ is its segmentation mask, $\mathbf T_{t,j}\in\mathrm{SE}(3)$ denotes its pose when applicable, $\mathbf b_{t,j}$ describes its geometric extent, and $z_{t,j}$ denotes its inferred task-relevant state.

The state descriptor $z_{t,j}$ converts geometric observations into task-relevant object states used by subsequent reasoning. For brick construction, $z_{t,j}$ encodes the predefined states flat, side-oriented, vertical, and stacked, as illustrated in \cref{fig:detailed_architecture}. The state descriptors provide a common interface across the evaluated tasks, while the instantiated state information depends on the physical interaction required by each task. 

Pick-and-place represents object and region identities, object poses, and their spatial relations; pyramid stacking additionally represents support and relative-placement relations; and clean board represents the cleaning tool, board plane, and detected residual region. The structured output of this module is passed to the subsequent language-grounding and action-generation module. Object identities are maintained across execution steps for correspondence, while masks, poses, state descriptors, and spatial relations are updated from the latest RGB-D observation. 

\begin{table*}[!t]
\centering
\caption{Task-dependent action-sequence composition, feasibility checks,
and execution verification using the shared atomic action library.}
\label{tab:action_sequences}
\begingroup
\footnotesize
\setlength{\tabcolsep}{3pt}
\renewcommand{\arraystretch}{1.06}

\begin{tabular}{p{2.4cm} p{4.2cm} p{5.4cm} p{4.8cm}}
\toprule
\textbf{Task} &
\textbf{Grounded objective} &
\textbf{Action-sequence composition} &
\textbf{Feasibility and verification conditions} \\
\midrule

Pick-and-Place &
Transfer the commanded object to the grounded target region or pose &
\textsc{Move} (above object) $\rightarrow$
\textsc{Descend} $\rightarrow$
\textsc{Close} $\rightarrow$
\textsc{Lift} $\rightarrow$
\textsc{Move} (target pose) $\rightarrow$
\textsc{Descend} $\rightarrow$
\textsc{Release} &
Reachability, collision, grasp success, and final pose or containment
\\

\addlinespace[3pt]

Pyramid Stacking &
Place each object at its grounded target while satisfying the required
support and relative-pose relations &
Shared pick-and-place sequence with state-conditioned \textsc{Rotate}
and additional \textsc{Move} actions for reorientation and separation &
Reachability, collision/contact, alignment, support relation, settling,
and final placement
\\

\addlinespace[3pt]

Clean Board &
Remove the grounded residual region while maintaining the required
interaction with the board &
\textsc{Move} (approach) $\rightarrow$
\textsc{Descend} $\rightarrow$
repeated contact-constrained \textsc{Move} $\rightarrow$
\textsc{Lift} &
Reachability, collision, contact condition, board boundary, residual
mask, and coverage
\\

\bottomrule
\end{tabular}

\endgroup
\end{table*}

\subsection{Language Grounding and Action-Sequence Generation}
\label{subsec:action-generation}

The cerebellum module converts the language command and perceived object
states into candidate action sequences. A central design of \method{} is
to avoid generating a long-horizon manipulation plan as a single
monolithic output. Long-horizon robot tasks can be decomposed into
atomic actions that are reused across different task
objectives, object configurations, and environments. We therefore
abstract these recurring operations from complex task executions into a
shared atomic action library. Rather than generating a complete
action sequence directly from the language command, \method{} separates
task grounding from action composition through an explicit grounded-task
representation, following the general principle of staged intermediate
reasoning \cite{wei2022cot}. As illustrated in
\cref{fig:detailed_architecture}, Stage~1 grounds the language command in
the perceived physical state, and Stage~2 composes atomic actions into
candidate action sequences.

The task-grounding stage receives the language command $L$, the current
scene state $S_t$, and the verified execution record $M_t$. The latter
stores task progress confirmed by previous execution and verification
steps. The grounded task is represented as
\begin{equation}
G_t
=
\Gamma(L,S_t,M_t)
=
\{g_{t,k}\}_{k=1}^{K_t}.
\label{eq:task_grounding}
\end{equation}

Each grounded goal is defined as
\begin{equation}
g_{t,k}
=
(\omega_k,\tau_k,\epsilon_k,v_k),
\label{eq:grounded_goal}
\end{equation}
where $\omega_k$ identifies the object or region referred to by the
command, $\tau_k$ specifies its desired physical state or spatial
relation, $\epsilon_k$ defines the admissible tolerance, and $v_k$
defines the verification condition used to determine whether the
grounded objective has been physically achieved. The grounded task
therefore specifies the required physical outcome without prescribing a
fixed action sequence.

For Pyramid Stacking, the grounding stage resolves the object to
manipulate, its target placement, and the required support and
relative-pose relations. For Pick-and-Place, it resolves the commanded
object and destination region or pose. For Clean Board, it associates
the cleaning command with the cleaning tool, board surface, and residual
region obtained from the visual-perception module. The output of Stage~1
is therefore a structured task description grounded in the current
physical state rather than free-form action text.

Stage~2 performs action composition using a shared atomic action library.
In the current implementation, the library $\mathcal{A}_{\mathrm{atom}}$ contains the following core
atomic actions:
\begin{itemize}
    \item \textsc{Move}: move the end effector toward a parameterized target pose or along a specified path;
    \item \textsc{Descend}: move the end effector along the task-defined approach or contact direction;
    \item \textsc{Rotate}: change the orientation of the end effector or manipulated object to a specified target orientation;
    \item \textsc{Close}: close the gripper according to the grounded grasp configuration;
    \item \textsc{Lift}: raise the grasped object to a specified clearance or transport height;
    \item \textsc{Release}: open the gripper to release the manipulated object at the target state.
\end{itemize}

Each atomic action represents a reusable robot elementary operation with explicit
geometric or interaction parameters. These actions are defined above the
low-level controller: trajectory interpolation, joint-level tracking, and
high-rate servo control remain handled by the robot controller. The same
atomic action library is shared across all three task families, while task
differences are expressed through action selection, ordering,
parameterization, and verification conditions. \Cref{tab:action_sequences}
summarizes how the shared atomic actions are composed and verified for the
three task families. The atomic action library is extensible rather than fixed to these six operations. New atomic actions can be incorporated by following the same
parameterization and verification interface, allowing the framework to
support additional manipulation tasks without changing the overall
hierarchical architecture.

Conditioned on the current physical state and grounded task, the
action-composition stage generates a set of candidate action sequences:
\begin{equation}
\mathcal{C}_t
=
\Psi(S_t,G_t,M_t;\mathcal{A}_{\mathrm{atom}})
=
\{\mathbf{a}_t^{(i)}\}_{i=1}^{N_t}.
\label{eq:candidate_sequences}
\end{equation}

The $i$-th candidate sequence is
\begin{equation}
\mathbf{a}_t^{(i)}
=
(a_{t,1}^{(i)},\ldots,a_{t,H_i}^{(i)}),
\label{eq:action_sequence}
\end{equation}
where $H_i$ denotes its sequence length. Each atomic action in the
sequence is represented as
\begin{equation}
a_{t,h}^{(i)}
=
(\rho,\omega,\xi,\kappa,\nu),
\label{eq:atomic_action}
\end{equation}
where $\rho \in \mathcal{A}_{\mathrm{atom}}$ denotes the action type,
$\omega$ specifies the manipulated object or interaction target, $\xi$
contains the geometric target parameters, $\kappa$ specifies the gripper
or contact parameters, and $\nu$ specifies the verification condition
associated with the corresponding execution stage.

The atomic action vocabulary is deliberately compact. Task-dependent
operations are obtained by parameterizing the shared actions rather than
introducing new primitives. In Pick-and-Place, \textsc{MoveAbove} and
\textsc{MoveToTarget} are parameterized instances of \textsc{Move}. A
nominal sequence is
\begin{equation}
\begin{aligned}
&\textsc{MoveAbove}
\rightarrow \textsc{Descend}
\rightarrow \textsc{Close} \rightarrow  \textsc{Lift} \\
& \rightarrow \textsc{MoveToTarget}
\rightarrow \textsc{Descend} \rightarrow \textsc{Release}.
\end{aligned}
\label{eq:pickplace_sequence}
\end{equation}
The target poses, approach directions, descent distances, and gripper
parameters are instantiated from $S_t$ and $G_t$.

Pyramid Stacking uses the same atomic actions, while the inferred object
state and grounded support relation determine their composition. A flat
object can proceed directly to grasping and placement, whereas a
side-oriented or vertical object requires \textsc{Rotate} or additional
parameterized \textsc{Move} actions before transport. A stacked object
requires an additional motion to separate it from its supporting object
before grasping. The generated sequence is therefore conditioned on the
observed physical state rather than fixed for the task category.

Clean Board also uses the shared atomic action library. The grounded
board plane and residual region determine the approach pose and planar
motion trajectory. Parameterized \textsc{Move} and \textsc{Descend}
actions establish the required interaction with the board, repeated
planar \textsc{Move} actions cover the residual region, and
\textsc{Lift} terminates the current wiping operation. After execution,
the updated residual region determines the subsequent grounded task
and action sequence.

Multiple candidate sequences can be generated for the same grounded task
when different object assignments, action orderings, geometric
parameters, or intermediate motions satisfy the required physical
outcome. The cerebellum module retains these alternatives rather than
selecting one according to language reasoning alone. The candidate set
$\mathcal{C}_t$ is passed to the prefrontal-cortex module in the next subsection, where physical
feasibility and sequence cost are evaluated before the task plan is
selected.

\subsection{Cost-Based Plan Selection}
\label{subsec:decision-optimization}

The prefrontal-cortex module selects a task plan from the candidate action
sequences generated by the cerebellum. As illustrated in
\cref{fig:detailed_architecture}, the module consists of three stages:
per-action cost evaluation, sequence-cost aggregation, and ranking and
selection. The underlying intuition is that, when multiple action
sequences can accomplish the same grounded task, the robot should prefer
the sequence that requires less motion and fewer additional operations.

For each atomic action $a_{t,h}^{(i)}$ in the $i$-th candidate sequence,
we define an execution-time cost as
\begin{equation}
c_{t,h}^{(i)}
=
\alpha_d d_{t,h}^{(i)}
+
\alpha_{\theta} \theta_{t,h}^{(i)}
+
\alpha_g g_{t,h}^{(i)},
\label{eq:action_cost}
\end{equation}
where $d_{t,h}^{(i)}$ is the translational displacement of the robot
end effector, $\theta_{t,h}^{(i)}$ is the absolute yaw rotation,
and $g_{t,h}^{(i)}$ indicates the number of gripper opening or closing operation. The coefficients convert the three components into a common time unit.
Specifically,
\begin{equation}
\alpha_d = \frac{1}{\bar{v}}, \qquad
\alpha_{\theta} = \frac{1}{\bar{\omega}}, \qquad
\alpha_g = \bar{t}_g,
\label{eq:cost_coefficients}
\end{equation}
where $\bar{v}$ is the measured average translational speed of the end
effector, $\bar{\omega}$ is the measured average yaw angular speed, and
$\bar{t}_g$ is the average time required for one gripper opening or
closing operation. These quantities are measured from repeated real-robot
executions and are fixed for all experiments.
The resulting cost therefore provides an estimate of the execution time
of each atomic action.

The cost of the complete $i$-th candidate action sequence is obtained by
summing the estimated execution times of its atomic actions:
\begin{equation}
J_t^{(i)}
=
\sum_{h=1}^{H_i}
c_{t,h}^{(i)},
\label{eq:sequence_cost}
\end{equation}
where $H_i$ denotes the number of atomic actions in the sequence. Longer
motions, larger reorientations, and additional atomic actions therefore
increase the accumulated sequence cost naturally through their estimated
execution times.

The candidate action sequences are ranked according to $J_t^{(i)}$, and
the selected task plan is
\begin{equation}
\mathbf{a}_t^{*}
=
\arg\min_{\mathbf{a}_t^{(i)} \in \mathcal{C}_t}
J_t^{(i)}.
\label{eq:plan_selection}
\end{equation}
When multiple action sequences can accomplish the same grounded task, this
cost provides a common physical criterion for selecting the sequence with
the lowest estimated execution time. The selected sequence is then passed
to the real-robot execution and verification module.

\subsection{Real-Robot Execution and Verification}
\label{subsec:execution-replanning}

The selected action sequence is executed on the robot and verified using
refreshed physical observations, as illustrated at the bottom of
\cref{fig:detailed_architecture}. The prefrontal-cortex module selects the
action sequence, while this module determines whether each executed atomic
action has achieved its intended physical outcome.

Let $t$ denote the current planning cycle and $h$ the index of an atomic
action within the selected sequence. The robot controller executes
$a_{t,h}^{*}$ according to
\begin{equation}
(q_{t,h+1},r_{t,h})
=
\mathcal{E}(a_{t,h}^{*},q_{t,h}),
\label{eq:robot_execution}
\end{equation}
where $q_{t,h}$ and $q_{t,h+1}$ are the robot states before and after
execution, $r_{t,h}$ is the controller status, and $\mathcal{E}$ denotes
the robot execution interface. Low-level trajectory interpolation and
servo control are handled by the robot controller.

After execution, a new RGB-D observation $O_{t,h+1}$ is acquired and the
scene state is refreshed using the perception function $\Phi$ defined in
\cref{subsec:perception-goals}:
\begin{equation}
S_{t,h+1}
=
\Phi(O_{t,h+1},q_{t,h+1}),
\label{eq:state_refresh}
\end{equation}
where $S_{t,h+1}$ represents the observed physical state after executing
$a_{t,h}^{*}$.

Each atomic action is associated with a set of verification conditions.
Let $\mathcal{K}_{t,h}$ denote the indices of the conditions associated
with $a_{t,h}^{*}$. The stage-level verification result is
\begin{equation}
\chi_{t,h}
=
\prod_{k\in\mathcal{K}_{t,h}}
v_k(S_{t,h+1},G_t),
\label{eq:stage_verification}
\end{equation}
where $v_k(\cdot)\in\{0,1\}$ is the $k$-th verification function defined
by the grounded task $G_t$, and $\chi_{t,h}=1$ only when all conditions
for the current atomic action are satisfied. The task-dependent
verification conditions are summarized in \cref{tab:action_sequences}.

If $\chi_{t,h}=1$, execution proceeds to the next atomic action. Otherwise,
the remaining sequence is discarded and a new action sequence is generated
from the refreshed scene state while previously verified task progress is
retained. The task terminates when all terminal verification conditions in
$G_t$ are satisfied. Thus, task completion is determined by observed
physical outcomes rather than by the generated action sequence alone.

\begin{table}[!t]
\centering
\caption{Task-progress definitions and terminal success criteria.}
\label{tab:task_definitions}
\begingroup
\scriptsize
\setlength{\tabcolsep}{2pt}
\renewcommand{\arraystretch}{0.96}
\begin{tabular}{p{1.6cm} p{3.6cm} p{2.2cm}}
\toprule
\textbf{Task} &
\textbf{Task progress $\mathrm{TP}_r$} &
\textbf{Success criterion} \\
\midrule

Pick-and-Place &
Verified blocks transferred to the bowl / required blocks &
$\mathrm{TP}_r=100\%$
\\

Pyramid Stacking &
Verified bricks correctly placed / target bricks &
$\mathrm{TP}_r=100\%$ and stable final structure
\\

Clean Board &
$100(1-|D_T|/|D_0|)$ &
$\mathrm{TP}_r>80\%$
\\

\bottomrule
\end{tabular}
\endgroup
\end{table}

\begin{table*}[!t]
\centering
\caption{Comparison with ReKep, Dream2Flow, and $\pi_{0.5}$ across the
three task families. SR reports success rate and successful trials;
completion time is computed over successful trials only; TP reports
mean $\pm$ sample standard deviation over all trials.}
\label{tab:baseline_compare}
\begingroup
\scriptsize
\setlength{\tabcolsep}{3pt}
\renewcommand{\arraystretch}{0.90}

\begin{tabular}{llccc}
\toprule
\textbf{Condition} &
\textbf{Method} &
\textbf{SR (\%; $n_s/N$)} &
\textbf{Time (s)} &
\textbf{TP (\%)} \\
\midrule

\multirow{4}{*}{Clean Board}
& ReKep & 0; 0/10 & -- & 15.22 $\pm$ 19.35 \\
& Dream2Flow & 10; 1/10 & 93.0 & 52.93 $\pm$ 26.50 \\
& $\pi_{0.5}$ & 30; 3/10 & 124.0 $\pm$ 9.67 & 42.46 $\pm$ 42.27 \\
& \method{} (ours) & \textbf{100; 10/10} &
171.3 $\pm$ 28.65 & \textbf{99.03 $\pm$ 1.67} \\

\midrule
\multirow{4}{*}{Pick-and-Place}
& ReKep & 10; 1/10 & 279.0 & 53.33 $\pm$ 28.11 \\
& Dream2Flow & 0; 0/10 & -- & 53.33 $\pm$ 17.21 \\
& $\pi_{0.5}$ & 50; 5/10 & 95.4 $\pm$ 40.66 & 70.00 $\pm$ 36.68 \\
& \method{} (ours) & \textbf{100; 10/10} &
182.1 $\pm$ 17.03 & \textbf{100.00 $\pm$ 0.00} \\

\midrule
\multirow{4}{*}{Pyramid--Flat}
& ReKep & 0; 0/5 & -- & 53.33 $\pm$ 32.06 \\
& Dream2Flow & 0; 0/5 & -- & 0.00 $\pm$ 0.00 \\
& $\pi_{0.5}$ & 40; 2/5 & 136.5 $\pm$ 14.85 & 73.33 $\pm$ 25.28 \\
& \method{} (ours) & \textbf{80; 4/5} &
306.0 $\pm$ 18.49 & \textbf{96.67 $\pm$ 7.45} \\

\midrule
\multirow{4}{*}{Pyramid--Irregular}
& ReKep & 0; 0/5 & -- & 50.00 $\pm$ 33.33 \\
& Dream2Flow & 0; 0/5 & -- & 0.00 $\pm$ 0.00 \\
& $\pi_{0.5}$ & 0; 0/5 & -- & 26.67 $\pm$ 14.91 \\
& \method{} (ours) & \textbf{80; 4/5} &
354.75 $\pm$ 42.52 & \textbf{96.67 $\pm$ 7.45} \\

\bottomrule
\end{tabular}

\endgroup
\end{table*}

\begin{figure*}[!t]
\centering
\includegraphics[width=0.9\linewidth]{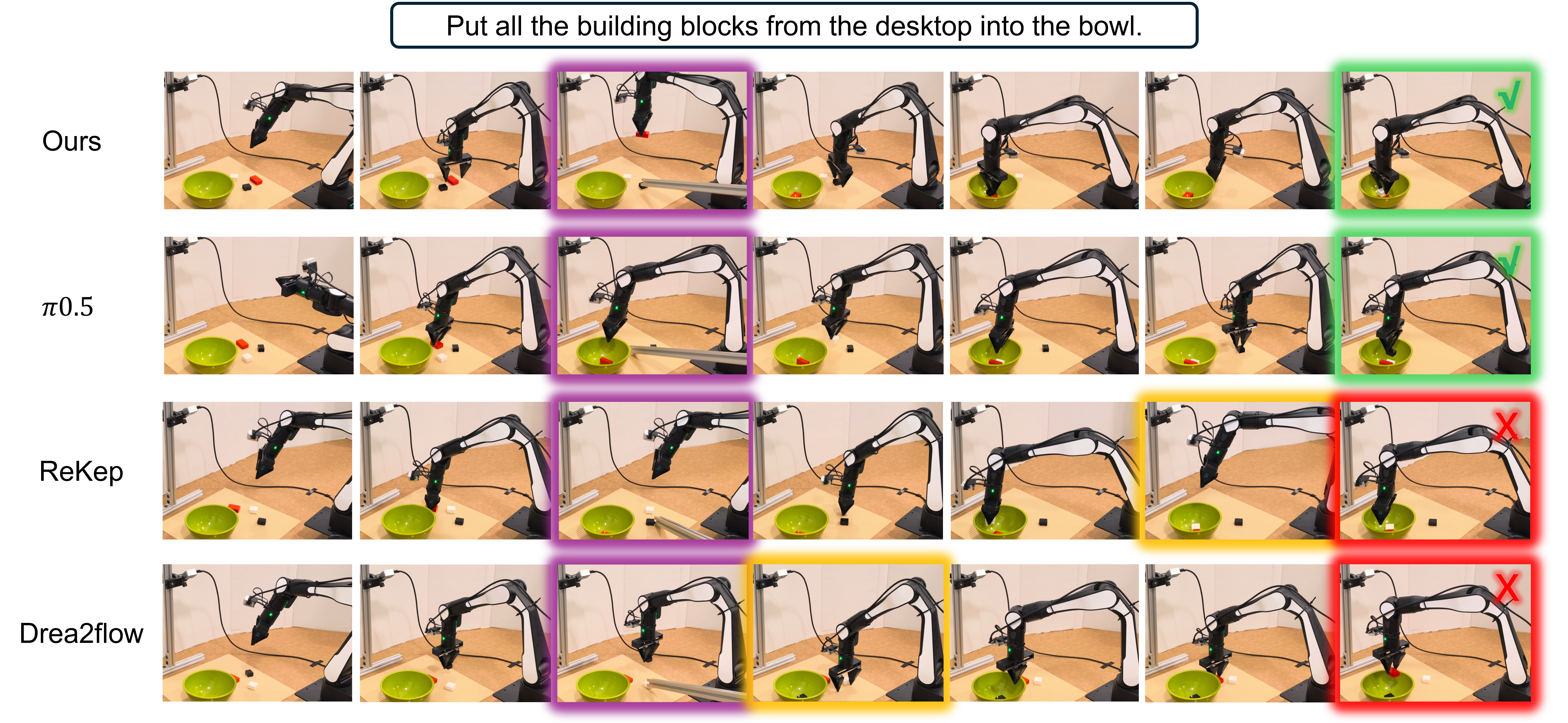}
\caption{Selected Pick-and-Place rollouts (rows: \method{}, $\pi_{0.5}$,
ReKep, and Dream2Flow). Frames progress from left to right. Purple 
highlights an externally introduced human disturbance, yellow a
step-level execution error, green task completion, and red an incomplete
terminal outcome.}
\label{fig:pickplace_qualitative}
\end{figure*}

\section{Experiments}
\label{sec:experiments}

\subsection{Experimental Setup}
\label{subsec:task-suite}

We evaluate \method{} on three complementary task families: repeated
Pick-and-Place, support-sensitive Pyramid Stacking under flat and irregular
initial layouts, and continuous-contact Clean Board. \emph{Zero-shot}
denotes execution without task-specific weight updates or manually
specified step-by-step action sequences. Across all tasks, the hierarchical reasoning pipeline, shared atomic action library, module interfaces, qwen3.7 plus backbone \cite{yang2025qwen3}, and low-level robot controller remain unchanged. Qwen3-7B is used for visual state reasoning, task grounding, and action composition, while task goals, action-sequence parameterization, and verification conditions are adapted to the physical requirements of each task. The execution-time coefficients used for
cost-based plan selection are calibrated from repeated real-robot executions and fixed across all experiments:
$\alpha_d=27.39~\mathrm{s/m}$ for translational motion, $\alpha_{\theta}=2.72~\mathrm{s/rad}$ for yaw rotation, and 
$\alpha_g=1.23~\mathrm{s}$ for each gripper opening or closing operation. For each task, we collected 50 episodes to fine-tune $\pi_{0.5}$.

The three task families evaluate different forms of long-horizon robot
execution. Pick-and-Place requires the robot to repeatedly identify,
grasp, transport, and place all commanded blocks into a bowl, with task
completion determined by final containment. Pyramid Stacking requires the
robot to construct the target structure from both flat and irregular
initial layouts, testing state-dependent action-sequence composition,
placement accuracy, and structural stability. Clean Board replaces
discrete object manipulation with continuous-contact wiping, where the
cleaning tool and marked board region are grounded from visual observations,
and the residual region determines the remaining task progress.


\subsection{Evaluation Protocol and Metrics}
\label{subsec:metrics}

Clean Board and Pick-and-Place use $N=\ntrials$ trials per method, while
each Pyramid Stacking condition uses $N=\npyramidtrials$. We evaluate
success rate (SR), task progress (TP), and completion time. Let
$s_r\in\{0,1\}$ denote the success indicator of trial $r$. The success
rate, reported as a percentage, is
\begin{equation}
\mathrm{SR}
=
\frac{100}{N}
\sum_{r=1}^{N}s_r.
\label{eq:success_rate}
\end{equation}

For Pick-and-Place and Pyramid Stacking, task progress is defined by the
fraction of required subgoals completed in trial $r$:
\begin{equation}
\mathrm{TP}_r
=
\frac{100}{K_r}
\sum_{k=1}^{K_r}z_{r,k},
\label{eq:task_progress_discrete}
\end{equation}
where $K_r$ is the number of required subgoals and
$z_{r,k}\in\{0,1\}$ indicates whether the $k$-th subgoal has been
completed and verified. A trial is successful when
$\mathrm{TP}_r=100\%$ and the terminal task conditions are satisfied.

For Clean Board, task progress is measured by the reduction of the marked
residual region:
\begin{equation}
\mathrm{TP}_r
=
100
\left(
1-\frac{|D_T|}{|D_0|}
\right),
\label{eq:task_progress_clean}
\end{equation}
where $D_0$ and $D_T$ denote the initial and final residual-region masks,
respectively. A Clean Board trial is considered successful when
$\mathrm{TP}_r>80\%$.

TP is reported as mean $\pm$ sample standard deviation over all trials.
Completion time is computed only over successful trials; ``--'' indicates
that a method produces no successful trial under the corresponding
condition. \Cref{tab:task_definitions} summarizes the task-progress
definitions and terminal success criteria.

\begin{figure*}[!t]
\centering
\includegraphics[width=0.9\linewidth]{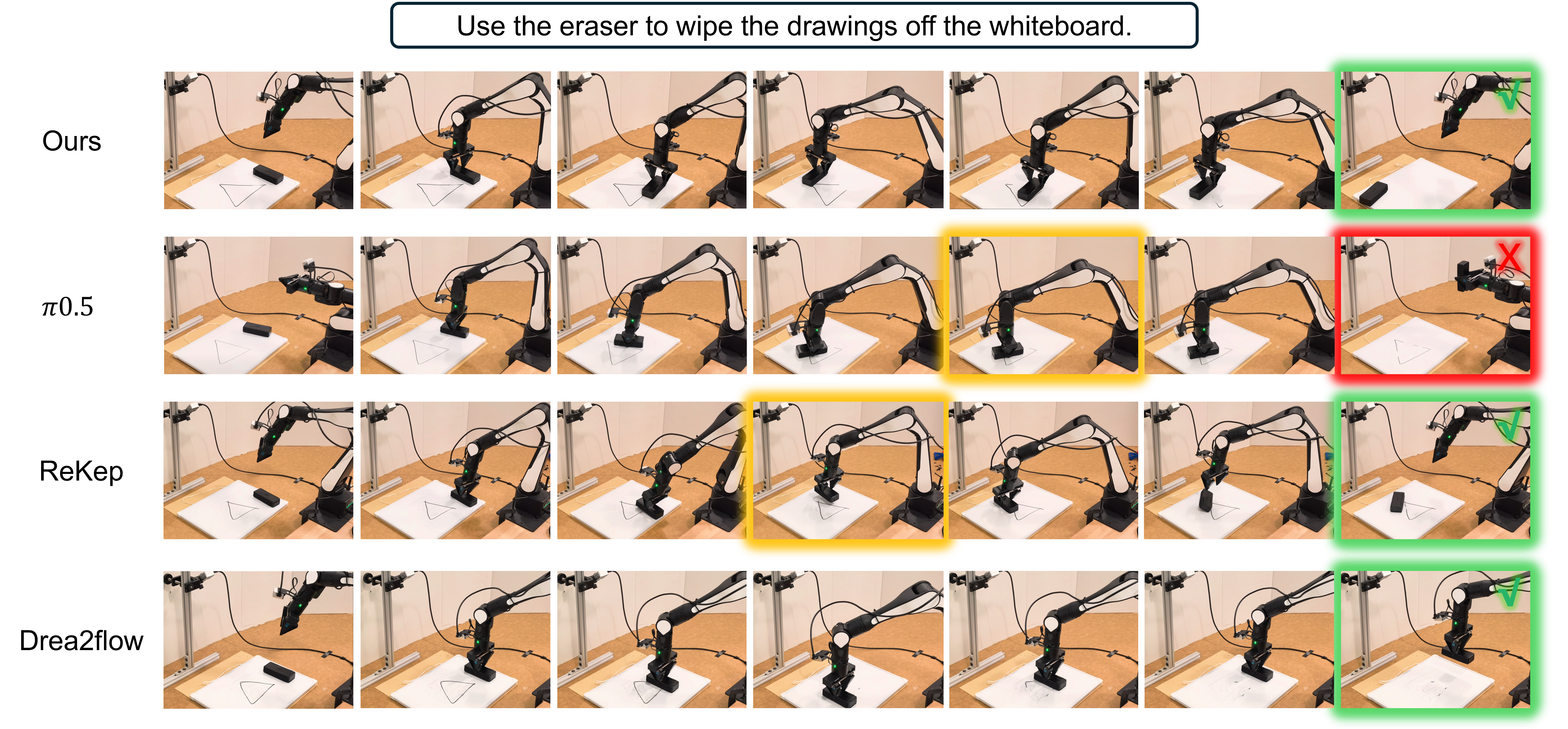}
\caption{Selected Clean Board rollouts in the same row order as
\cref{fig:pickplace_qualitative}. Yellow highlights the onset of
ineffective wiping caused by a stall or missed region, green task
completion, and red incomplete cleaning. Quantitative success is defined
by $\mathrm{TP}>80\%$.}
\label{fig:cleanboard_qualitative}
\end{figure*}

\begin{figure*}[!t]
\centering
\includegraphics[width=0.9\linewidth]{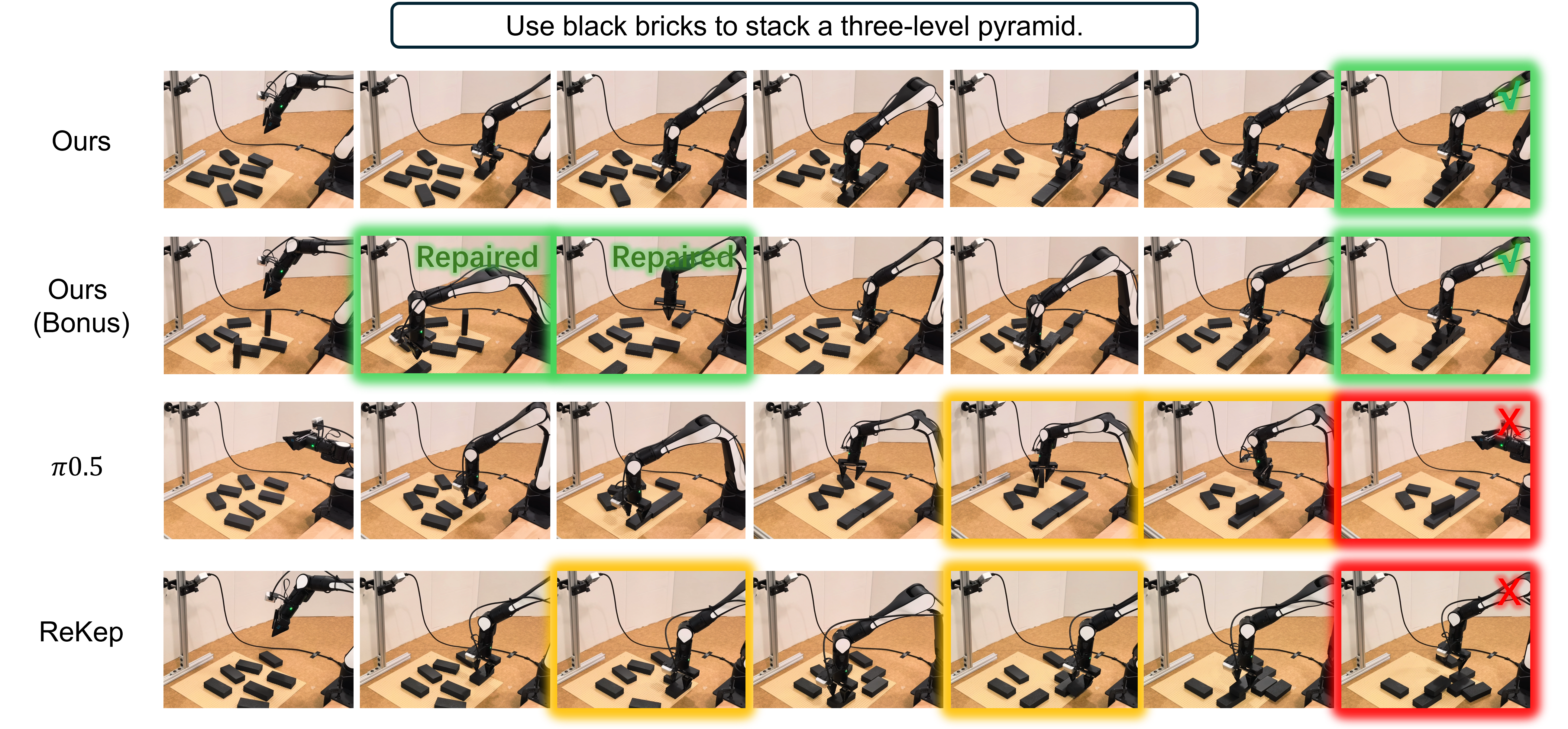}
\caption{Selected Pyramid Stacking rollouts for nominal and recovery
\method{} cases, $\pi_{0.5}$, and ReKep. Yellow highlights the onset of a
failed grasp, green task completion or repaired case, and red incomplete construction. The
repaired case shows \method{} reorienting an irregularly placed brick to a
flat state before continuing construction.}
\label{fig:pyramid_qualitative}
\end{figure*}

\subsection{Comparison With Baselines}
\label{subsec:baseline-results}

We compare \method{} with ReKep \cite{huang2024rekep}, Dream2Flow
\cite{dharmarajan2025dream2flow}, and $\pi_{0.5}$
\cite{black2025pi05} under the same task definitions and terminal success
criteria. As shown in \cref{tab:baseline_compare}, \method{} achieves the
highest success rate and task progress in all evaluated conditions,
including 100\% success on Clean Board and Pick-and-Place and 80\% success
under both Pyramid Stacking layouts. The performance gap is particularly
clear for Pyramid--Irregular, where none of the baselines completes a
trial while \method{} succeeds in 4/5 trials.

These results support the advantage of explicitly structuring long-horizon
robot execution. \method{} grounds language commands in the observed object
states, composes task-dependent action sequences from a shared atomic
action library, and selects among candidate sequences using execution cost.
Intermediate physical verification further allows the framework to update
task progress from updated observations rather than assuming that a
generated action has succeeded. A representative case is
Pyramid--Irregular: after a failed grasp leaves a brick in an irregular
orientation, \method{} re-infers its state and composes additional
\textsc{Rotate} and \textsc{Move} atomic actions before resuming the
nominal placement sequence, as illustrated in
\cref{fig:pyramid_qualitative}. This explicit intermediate-action
composition enables recovery from a state in which the original sequence
is no longer applicable. Completion time is reported only over successful
trials and is therefore treated as a secondary efficiency measure.

\subsection{Qualitative Results}
\label{subsec:qualitative-results}

\Cref{fig:pickplace_qualitative,fig:cleanboard_qualitative,fig:pyramid_qualitative}
show representative rollouts that complement the quantitative results in
\cref{tab:baseline_compare}. Frames progress from left to right, allowing
intermediate state changes, execution errors, and recovery behavior to be
observed directly.

In \cref{fig:pickplace_qualitative}, an external human disturbance changes
the object position and orientation during execution. \method{} updates the object
state and continues execution from the updated scene rather than relying
on the original action sequence. The rollout illustrates the role of
closed-loop perception and verification when the physical state changes
during a long-horizon task.

In \cref{fig:cleanboard_qualitative}, task completion depends on the
observed cleaning result rather than on completion of a nominal wiping
motion. \method{} updates the residual region after wiping and continues
with another action sequence when required. This demonstrates the benefit
of execution verification for continuous-contact tasks in which motion
completion does not necessarily imply task completion.

The recovery case in \cref{fig:pyramid_qualitative} further demonstrates
the interaction between the modules. After a failed grasp leaves a brick
in an irregular orientation, the new observation updates its object
state, the cerebellum composes a reorientation sequence from the shared
atomic actions, and execution resumes until the structure is completed.
This example highlights how object-state inference, state-conditioned
action-sequence generation, and intermediate verification together enable
recovery from execution errors in long-horizon construction.

\FloatBarrier
\begin{table}[!t]
\centering
\caption{Component ablations. TP reports mean $\pm$ sample standard deviation.}
\label{tab:component_ablation}
\begingroup
\scriptsize
\setlength{\tabcolsep}{1.2pt}
\renewcommand{\arraystretch}{0.80}
\begin{tabular}{@{}llcc@{}}
\toprule
\textbf{Condition} &
\textbf{Variant} &
\shortstack{\textbf{SR}\\[-1pt]\textbf{$n_s/N$ (\%)}} &
\shortstack{\textbf{TP}\\[-1pt]\textbf{(\%)}} \\
\midrule

\multirow{3}{*}{Clean Board}
& Full \method{} & \textbf{10/10 (100)} & \textbf{99.03 $\pm$ 1.67} \\
& Single agent & 8/10 (80) & 91.55 $\pm$ 16.38 \\
& Single LLM & 0/10 (0) & 0.00 $\pm$ 0.00 \\

\midrule
\multirow{3}{*}{Pick-and-Place}
& Full \method{} & \textbf{10/10 (100)} & \textbf{100.00 $\pm$ 0.00} \\
& Single agent & 7/10 (70) & 86.67 $\pm$ 23.31 \\
& Single LLM & 8/10 (80) & 93.33 $\pm$ 14.05 \\

\midrule
\multirow{5}{*}{Pyr.--Flat}
& Full \method{} & \textbf{4/5 (80)} & \textbf{96.67 $\pm$ 7.45} \\
& Single agent & 1/5 (20) & 50.00 $\pm$ 33.33 \\
& Single LLM & 0/5 (0) & 0.00 $\pm$ 0.00 \\
& No state inference & \textbf{4/5 (80)} & \textbf{96.67 $\pm$ 7.45} \\
& No cost-based selection & 2/5 (40) & 46.67 $\pm$ 49.16 \\

\midrule
\multirow{5}{*}{Pyr.--Irregular}
& Full \method{} & \textbf{4/5 (80)} & \textbf{96.67 $\pm$ 7.45} \\
& Single agent & 1/5 (20) & 50.00 $\pm$ 31.18 \\
& Single LLM & 0/5 (0) & 0.00 $\pm$ 0.00 \\
& No state inference & 0/5 (0) & 36.67 $\pm$ 7.45 \\
& No cost-based selection & 0/5 (0) & 30.00 $\pm$ 18.26 \\

\bottomrule
\end{tabular}
\endgroup
\end{table}

\subsection{Ablation Study}
\label{subsec:ablation-results}

\Cref{tab:component_ablation} evaluates four variants that remove or
simplify key parts of the proposed hierarchy: specialized reasoning roles,
staged LLM reasoning, visual state inference, and cost-based action-sequence
selection. All variants use the same task definitions and low-level robot
controller.

\noindent\textbf{{Single agent.}}
Merging the specialized reasoning roles reduces performance across all
three task families, with the largest degradation occurring in Pyramid
Stacking. The result suggests that separating task grounding, action
generation, and plan selection becomes increasingly important as execution
requires longer action sequences and more intermediate physical states.

\noindent\textbf{{Single LLM.}}
Using a single LLM call retains relatively strong performance on
Pick-and-Place but fails to complete Clean Board and Pyramid Stacking.
This contrast supports the staged design of \method{}, where an explicit
grounded-task representation separates language understanding from
action-sequence composition. A direct model call can recover a relatively
standard pick-and-place pattern, but is less reliable when execution
requires long-horizon composition or repeated state-dependent decisions.

\noindent\textbf{{No state inference.}}
Removing visual state inference has little effect under the flat Pyramid
layout but causes a substantial degradation under the irregular layout.
This result directly supports the role of the visual-cortex module:
explicitly inferring whether an object is flat, side-oriented, vertical,
or stacked becomes important when the observed configuration differs from
the nominal case and the subsequent action sequence must be adapted
accordingly.

\noindent\textbf{No cost-based selection.}
Removing the prefrontal-cortex selection module degrades both Pyramid
conditions, with a larger effect under the irregular layout. Without
per-action and sequence-level cost evaluation, the framework cannot
systematically prefer shorter and simpler action sequences when multiple
candidates satisfy the same grounded task. The result therefore supports
the proposed separation between candidate generation in the cerebellum
and cost-based ranking in the prefrontal cortex.


\raggedbottom
\section{Conclusion}
\label{sec:conclusion}

We presented \method{}, a zero-shot hierarchical framework functionally inspired by the division of roles in the human brain, connecting language-guided task understanding with physically grounded robot execution through visual state inference, language-grounded action-sequence generation from a shared atomic action library, cost-based plan selection, and execution verification in a closed loop. Experiments across Pick-and-Place, Pyramid Stacking, and Clean Board show that \method{} achieves the highest success rate and task progress in all evaluated conditions compared with ReKep, Dream2Flow, and $\pi_{0.5}$. The ablation studies further support the complementary roles of the hierarchy: visual state inference enables adaptation to irregular physical configurations, staged grounding and action composition support long-horizon execution, and cost-based selection improves the choice among alternative action sequences. More importantly, the modular pipeline is not tied to a specific manipulation task: its shared atomic action representation can be composed and parameterized for different task objectives, while task-specific physical requirements are incorporated through grounding and verification conditions, providing a general interface for extending the framework to broader robot tasks and embodiments. Future work will further improve reasoning efficiency, expand the atomic action library, and evaluate the framework across a wider range of manipulation scenarios and robotic platforms.

\bibliographystyle{IEEEtran}
\bibliography{main}

\end{document}